\documentclass[11pt,a4paper]{article}
\usepackage[hyperref]{acl2020}
\usepackage{times}
\usepackage{latexsym}
\usepackage{graphicx}

\usepackage{microtype}

\aclfinalcopy 
\def\aclpaperid{3318} 

\title{Learning to Understand Child-directed and Adult-directed Speech}

\author{Lieke Gelderloos \\
  Tilburg University \\
  \texttt{l.j.gelderloos@uvt.nl} \\\And
  Grzegorz Chrupa{\l}a \\
  Tilburg University \\
  \texttt{g.chrupala@uvt.nl} \\\And
  Afra Alishahi \\
  Tilburg University \\
  \texttt{a.alishahi@uvt.nl}}

\date{}

\begin{document}
	\maketitle
	
	\begin{abstract}
			Speech directed to children differs from adult-directed speech in linguistic aspects such as repetition, word choice, and sentence length, as well as in aspects of the speech signal itself, such as prosodic and phonemic variation. Human language acquisition research indicates that child-directed speech helps language learners. This study explores the effect of child-directed speech when learning to extract semantic information from speech directly. We compare the task performance of models trained on adult-directed speech (ADS) and child-directed speech (CDS). We find indications that CDS helps in the initial stages of learning, but eventually, models trained on ADS reach comparable task performance, and generalize better. The results suggest that this is at least partially due to linguistic rather than acoustic properties of the two registers, as we see the same pattern when looking at models trained on acoustically comparable synthetic speech. 
		
	\end{abstract}
	
	\section{Introduction}
	Speech directed to children (CDS) differs from adult-directed speech (ADS) in many aspects. Linguistic differences include the number of words per utterance, with utterances in CDS being considerably shorter than utterances in ADS, and repetition, which is more common in child-directed speech. There are also paralinguistic, acoustic factors that characterize child-directed speech: people speaking to children typically use a higher pitch and exaggerated intonation.
	
	It has been argued that the properties of CDS help perception or comprehension. \citet{kuhl1997cross} propose that CDS is optimized for \textit{learnability}. Optimal learnability may, but does not necessarily align with optimization for perception or comprehension. Although speech with lower variability may be easiest to learn to understand, higher variability may provide more learning opportunities, leading to more complete language knowledge. 
	
	In this paper, we explore how learning to extract meaning from speech differs when learning from CDS and ADS. We discuss task performance on the training register as well as generalization across registers. To tease apart the effect of acoustic and linguistic differences, we also report on models trained on synthesized speech, in which linguistic differences between the registers are retained, but the acoustic properties are similar.
	
	\section{Related work}
	
	\subsection{Child directed speech and learnability}
	The characteristics of child-directed speech are a major topic of study in language acquisition research. For a comprehensive overview, see \citet{soderstrom2007beyond} and \citet[Ch.\ 2, p.\ 32-41]{clark2009first}. With regards to acoustics, CDS is reported to have exaggerated intonation and a slower speech rate \citep{fernald1989cross}. \citet{kuhl1997cross}  show that CDS contains more `extreme' realizations of vowels. \citet{mcmurray2013infant} show that these increased means are accompanied by increased variance, and argue that any learning advantage of CDS due to extreme vowel realizations is counteracted by increased variance. However, it has also been argued that increased variance may be beneficial to learning in the long run, as it gives the learner a more complete set of examples for a category, which helps generalization. 
	\citet{guevara2018words} show that word forms in child-directed speech are acoustically more diverse. At the utterance level, child-directed language consists of shorter sentences and simpler syntax \citep{newport1977mother, fernald1989cross}, and words more often appear in isolation \citep{ratner2001accessible}. 
	
	Studies on home recordings show that the availability of CDS input accounts for differences in vocabulary growth between learners, whereas overheard speech is unrelated \citep{Hoff03thespecificity, weisleder2013talking}. This does not necessarily mean that it is easier to learn from CDS. Psycholinguistic research has shown that infants across the world show a CDS preference, paying more attention to it than to ADS \citep{manybabies2020quantifying}. Learning advantages of CDS in children may therefore simply be because they grant it more attention, rather than to properties of CDS that are advantageous for learning. Computational models, however, have no choice in where they allocate attention. Any learning advantages we find of either ADS or CDS in computational studies must be due to properties that make speech in that register more learnable to the model.
	
	There has been some computational work comparing learning from ADS and CDS at the level of word learning and phonetic learning. Studies on segmentability use algorithms that learn to identify word units, with some studies reporting higher segmentability for CDS \citep{batchelder2002bootstrapping, daland2011learning}, while \citet{cristia2019segmentability} report mixed results. \citet{kirchhoff2005statistical} train HMM-based speech recognition systems on CDS and ADS, and test on matched and crossed test sets. They find that both ADS and CDS trained systems perform best on the matching test set, but CDS trained systems perform better on ADS than systems trained on ADS peform on CDS. They show that this is likely caused by phonetic classes have larger overlaps in CDS.
	
	To the authors' knowledge, the current work is the first to computationally explore learnability differences between ADS and CDS considering the process of speech comprehension as a whole: from audio to semantic information.

	\subsection{Speech recognition with non-linguistic supervision}
	In recent years, several studies have worked on machine learning tasks in which models directly extract semantic information from speech, without feedback on the word, character, or phoneme level. Most prominently, work on `weakly supervised' speech recognition includes work in which accompanying visual information is used as a proxy for semantic information. By grounding speech in visual information accompanying it, models can learn to extract visually relevant semantic information from speech, without needing symbolic annotation \citep{harwath2016unsupervised, harwath2017learning, chrupala2017representations, merkx2019language}.
	
	The topic is of interest for automatic speech recognition, as it provides potential ways of training speech recognition without the need for vast amounts of annotation. The utilization of non-linguistic information as supervision is particularly useful for low-resource languages. For the purpose of this study, however, we are interested in this set of problems because of the parallel to human language acquisition. A language learning child does not receive explicit feedback on the words or phonemes it perceives. Rather, they learn to infer these structural properties of language, with at their disposal only the speech signal itself and its weak and messy links to the outer world.
	
	\section{Task}
	The task is to match speech to a semantic representation of the language it contains, intuitively `grounding' it to the semantic context. The design of this task is inspired by work in visual grounding. However, the availability of CDS data accompanied by visual data is very limited. Instead of visual representation, we use semantic sentence embeddings of the transcriptions. Rather than training our model to imagine the visual context accompanying an utterance, as in visual grounding, we train it to imagine the semantic content. Note that since the semantic embeddings are based on the transcriptions of the sentences themselves, they have a much closer relation to the sentences than visual context representations would have.
	
	The semantic sentence representations were obtained using SBERT, a BERT-based architecture that yields sentence embeddings, which was finetuned on the STS benchmark of SemEval \citep{reimers-2019-sentence-bert}. This particular encoding was chosen because it harnesses the semantic strength of BERT  \citep{devlin-etal-2019-bert} in an encoding of the sentence as a whole. Speech is converted Mel-frequency cepstrum coefficients. 
	
	\section{Data}
	\subsection{Natural speech: NewmanRatner corpus}
	Since we are interested in the effect of learning from child- versus adult directed speech, we select data that differs in register, but is otherwise as comparable as possible. The NewmanRatner corpus contains annotated recordings of caregivers in conversation with their children and with experimenters \cite{newman2016input}. This dataset is suitable to our set-up, as it contains a reasonable amount of transcribed CDS and ADS by the same speakers, which is rare; and it is in English, for which pretrained state-of-the-art language models such as (S)BERT \citep{devlin-etal-2019-bert, reimers-2019-sentence-bert} are readily available. 
	
	Child-directed speech in the NewmanRatner corpus takes place in free play between caregiver and child, whereas adult-directed speech is uttered in the context of an interview. Stretches of speech have been transcribed containing one or more utterances. We selected only utterances by caregivers and excluded segments with multiple speakers. As the CDS portion of the corpus is larger than the ADS portion, we randomly selected 21,465 CDS segments, matching the number of ADS segments by caregivers. Validation and test sets of 1,000 segments were held out, while the remaining 19,465 segments were used for training. 
	
	Table \ref{Tab:descriptives} lists some characteristic statistics of the CDS and ADS samples that were used. The ADS sample contains a larger vocabulary than the CDS sample. On average, ADS segments contain more than twice as many words, although they are only 88 milliseconds longer on average. Therefore, the number of words per second is twice as high in ADS as it is in CDS.

	\begin{table}[t]
	\begin{center}
	\begin{tabular}{lrr}\hline
	\textbf{Dataset} & \textbf{CDS} & \textbf{ADS} \\\hline
	Vocabulary size	& 3,170 & 5,665 \\
	Total nr. of words & 97,118 & 203,084 \\
	Type/token ratio & .033 & .028 \\
	Words per utterance	& 4.52 & 9.46 \\
	Utterance length in seconds & 3.37 & 3.46\\
	Words per second & 1.34 & 2.74 \\\hline
	\end{tabular}
	\end{center}
	\caption{Descriptive statistics of the data}\label{Tab:descriptives}
	\end{table}

	\subsection{Synthetic speech}
	To tease apart effects of the acoustic properties of speech and properties of the language itself, we repeat the experiment using synthesized version of the ADS and CDS corpora. For this variant, we feed the transcriptions to the Google text2speech API, using the 6 available US English WaveNet voices \citep{oord2016wavenet}. Note that the synthetic speech is much cleaner than the natural speech, which was recorded using a microphone attached to clothing of the caregiver, and contains a lot of silence, noise, and fluctuations in volume of the speech.
	
	Since synthetic speech for ADS and CDS is generated using the same pipeline, the acoustic properties of these samples are comparable, but linguistic differences between them are retained. Differences remain in the vocabulary size, number of words per utterance and type token ratio, but the number of words per second is now comparable. This means the length of utterances is much larger for synthetic ADS sentences, since the average ADS sentence contains approximately twice as many words as the average CDS sentence.
	
	\section{Model}
	The model and training set-up is based on \citet{merkx2019language}. This model is suited to our task, as it allows to learn to extract semantic information from speech  by grounding it in another modality, without requiring the speech to be segmented. The speech encoder comprises a convolutional filter over the speech input, feeding into a stack of 4 bidirectional-GRU layers followed by an attention operator. The difference in our set-up is the use of SBERT sentence embeddings instead of visual feature vectors. Using a margin loss, the model is trained to make the cosine distance between true pairs of speech segments and SBERT embeddings smaller than that between random counterparts. We train for 50 epochs and following \citet{merkx2019language} we use a cyclic learning rate schedule.\footnote{Code is available through Github:\newline \texttt{https://github.com/lgelderloos/cds\_ads}}
	
	\section{Results}
	\subsection{Performance}
		\begin{table}[t]
		\vspace{4pt}
		\begin{center}
		\begin{tabular}{rrrrr}\hline
			\multicolumn{5}{c}{\textbf{Model trained on CDS}}\\\hline
			Testset & Med.r. & R@1 & R@5 & R@10 \\
			CDS & 5.00 & .27 & .51 & .60 \\
			ADS & 59.00 & .07 & .18 & .24 \\
			Combined & 34.00 & .14 & .29 & .36\\\hline
			\multicolumn{5}{c}{\textbf{Model trained on ADS}}\\\hline
			Testset & Med.r. & R@1 & R@5 & R@10 \\
			CDS & 45.67 & .09 & .22 & .30 \\
			ADS & 6.00 & .28 & .49 & .59 \\
			Combined & 24.67 & .15 & .31 & .39 \\\hline
		\end{tabular}
		\caption{Test performance of models trained on natural speech} 
		\label{Tab:test_natural}
		\vspace{12pt}
		\begin{tabular}{rrrrr}\hline
			\multicolumn{5}{c}{\textbf{Model trained on synthetic CDS}}\\\hline
			Testset & Med.r. & R@1 & R@5 & R@10 \\
			CDS & 1.00 & .82 & .96 & .99 \\
			ADS & 1.00 & .57 & .76 & .84 \\
			Combined & 1.00 & .66 & .83 & .88 \\\hline
			\multicolumn{5}{c}{\textbf{{Model trained on synthetic ADS}}}\\\hline	
			Testset & Med.r. & R@1 & R@5 & R@10 \\
			CDS & 1.00 & .68 & .88 & .94 \\
			ADS & 1.00 & .83 & .93 & .96 \\
			Combined & 1.00 & .72 & .88 & .92 \\\hline
		\end{tabular}
		\caption{Test performance of models trained on synthetic speech}
		\label{Tab:test_synth}
		\end{center}
	\end{table}
	Trained models are evaluated by ranking all SBERT embeddings in the test set by cosine distance to speech encodings. Reported metrics are recall@1, recall@5, and recall@10, which are the proportion of cases in which the correct SBERT embedding is among the top 1, 5, or 10 most similar ones; and the median rank of the correct SBERT embedding. Test results are reported for the training epoch for which recall@1 is highest on validation data. We have trained 3 differently randomly initialized runs for all four datasets, and report the average scores on the test split of the dataset the model was trained on, as well as its CDS or ADS counterpart, and a combined test set, which is simply the union of the two.
		
	As can be observed in table \ref{Tab:test_natural}, on the combined test set, models trained on adult directed speech slightly outperform models trained on child-directed speech. However, models in the two registers perform very similarly when we test them on the test set in the same register, with ADS having higher recall@1, but CDS scoring better on the other metrics. When we test ADS models on CDS, performance is lower than that of models that have been trained on CDS. However, the drop on ADS between models trained on ADS and models trained on CDS is even larger. The better performance on the combined test set, then, seems to come from ADS models generalizing better to CDS than the other way around.
	
	General performance of all models trained and tested on synthetic speech, which is much cleaner than the natural speech and more similar across registers, is much higher than performance on natural speech (see table \ref{Tab:test_synth}). However, the same pattern can be observed: on the combined test set, ADS models perform better than CDS models. When tested on the register they were trained on, the models perform similarly, but models trained on ADS perform better when tested on CDS than the other way around. 
	
	To summarize, models trained on ADS and CDS reach comparable scores when evaluated on the same register they are trained on. However, training on ADS leads to knowledge that generalizes better than training on CDS does. This pattern holds even when training and evaluating on synthetic speech, when the two registers are acoustically similar.

	\subsection{Learning trajectories}
	
	\begin{figure}[t]
		\includegraphics[width=\linewidth]{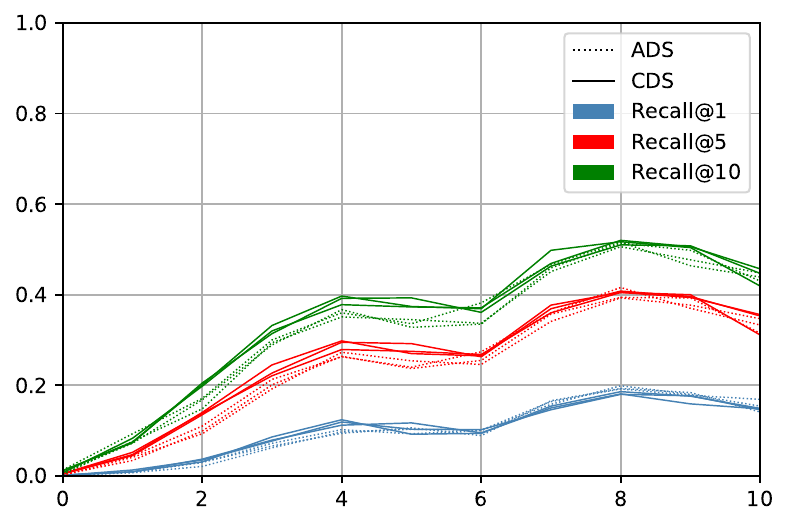}
		\caption{Validation performance in early training on natural speech}
		\label{fig:Trajectories_10_n}
		\vspace{4pt}
		\includegraphics[width=\linewidth]{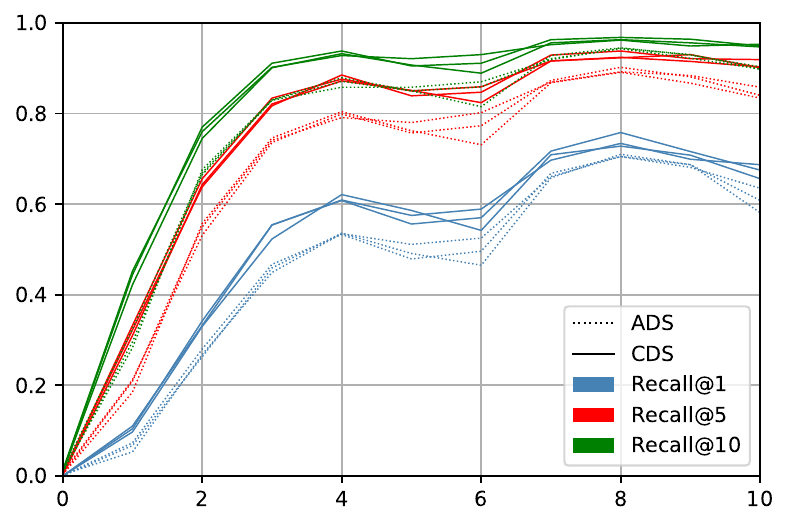}
		\caption{Validation performance in early training on synthetic speech}
		\label{fig:Trajectories_10_s}
	\end{figure}
	Learnability is not just about eventual attainment: it is also about the process of learning itself. Although ADS and CDS models eventually perform similarly, this is not necessarily the case during the training process. Figures \ref{fig:Trajectories_10_n} and \ref{fig:Trajectories_10_s} show the trajectory of recall performance on the validation set after the first 10 epochs of training. During the early stages of learning, the models trained on ADS (dotted lines) are outperformed by those trained on CDS (solid lines). This pattern is more pronounced in the models trained on synthetic speech, but also present for models trained on natural speech. After four epochs of training, average  recall@1 is 0.10 for ADS models and 0.12 for CDS models. For models trained on synthetic speech, average recall@1 on validation data is 0.53 for ADS models and 0.61 for CDS models. In later stages of training, models trained on ADS outperform CDS models on validation data. At epoch 40, close to the optimally performing epoch for most models, average recall@1 is 0.29 for ADS models and 0.27 for CDS models, and 0.85 and 0.81 for the synthetic counterparts, respectively. 
	
	Although models trained on adult-directed speech eventually catch up with models trained on child-directed speech, CDS models learn more quickly at the start.

	\section{Discussion}
	
	We find indications that learning to extract meaning from speech is initially faster when learning from child-directed speech, but learning from adult-directed speech eventually leads to  similar task performance on the training register, and better generalization to the other register. The effect is present both in models trained on natural speech and in models trained on synthetic speech, suggesting that it is at least partly due to differences in the language itself, rather than acoustic properties of the speech register.
	
	Our finding that models trained on ADS generalize better to CDS than the other way around contrasts with the findings of \citet{kirchhoff2005statistical}. Our results are in contrast to the idea that CDS is optimized for leading to the most valuable knowledge, as it is the models trained on ADS that lead to better generalization. Our finding that learning is initially faster for CDS is more in line with the idea of learnability as `easy to learn'.
	
	The better generalization of models trained on ADS may be due to ADS having higher lexical and semantic variability, reflected in the larger vocabulary and higher number of words per utterance. Since there is simply more to learn, learning to perform the task is more difficult on ADS, but it leads to more valuable knowledge. It is also possible that SBERT is better suited to encode the semantic content of ADS, as ADS utterances are likely to be more similar to the sentences SBERT was trained on than CDS utterances are.
	
	We must be prudent in drawing conclusions from the apparent effects we see in this study, as the results on different datasets cannot be interpreted as being on the same scale. Although all metrics are based on a rank of the same number of competitors, the distribution of similarities and differences between the semantic representations of these competitors may differ across datasets. The combined test set scores are more directly comparable, but ideally, we would like to compare the generalization of both models on an independent test set.
	
	In future work, we intend to curate a test set with data from separate sources, which can serve as a benchmark for the models we study. We intend to explore how a curriculum of CDS followed by ADS affects learning trajectories and outcomes. We also intend to use tools for interpreting the knowledge encoded in neural networks (such as diagnostic classifiers and representational similarity analysis) to investigate the emergent representation of linguistic units such as phonemes and words.

	\bibliography{biblio}
	\bibliographystyle{acl_natbib}
	
	\appendix
	
\end{document}